\pdfoutput=1

\documentclass[11pt]{article}

\usepackage[]{ACL2023}

\usepackage{times}
\usepackage{latexsym}
\usepackage{booktabs}
\usepackage{lipsum}
\usepackage{graphicx}
\usepackage{float}
\usepackage[breaklinks]{hyperref}
\usepackage{adjustbox}
\usepackage[T1]{fontenc}

\usepackage[utf8]{inputenc}

\usepackage{microtype}

\usepackage{inconsolata}

\title{It’s not Sexually Suggestive; It’s Educative | Separating Sex Education from Suggestive Content on TikTok videos 
}
\newcommand\tab[1][1cm]{\hspace*{#1}}

 \author{Enfa George \tab Mihai Surdeanu \\
         University Of Arizona  \\ {\{enfageorge,msurdeanu\}}@arizona.edu}

\begin{document}
\maketitle
\begin{abstract}
We introduce SexTok, a multi-modal dataset composed of TikTok videos labeled as sexually suggestive (from the annotator's point of view), sex-educational content, or neither. Such a dataset is necessary to address the challenge of distinguishing between sexually suggestive content and virtual sex education videos on TikTok. Children’s exposure to sexually suggestive videos has been shown to have adversarial effects on their development \citep{collins2017sexual}. Meanwhile, virtual sex education, especially on subjects that are more relevant to the LGBTQIA+ community, is very valuable \citep{mitchell2014accessing}. The platform's current system removes/punishes some of both types of videos, even though they serve different purposes. Our dataset contains video URLs, and it is also audio transcribed. To validate its importance, we explore two transformer-based models for classifying the videos. Our preliminary results suggest that the task of distinguishing between these types of videos is learnable but challenging. These experiments suggest that this dataset is meaningful and invites further study on the subject.

\end{abstract}

\section{Introduction}

In short-form videos such as in TikTok, accurately identifying sexually suggestive and sex education content amidst a sea of diverse video types poses a significant challenge.  In this paper, we delve into this problem, focusing specifically on TikTok, the most downloaded app in 2022, which has a sub\- stantial user base of early adolescents and young individuals (10-19: 32.5\%, 20-29: 29.5\%) \footnote{\url{https://wallaroomedia.com/blog/social-media/tiktok-statistics/}} 

The distinction between suggestive videos and virtual sex education holds crucial significance on multiple fronts. Adolescent sex education in the United States is delivered in a fragmented and often inadequate system, which has long been the subject of intense criticism and is vulnerable to political influence \cite{fowler2021let}.In this context, TikTok presents a novel and promising avenue to conveying comprehensive and accessible sexual health information to adolescents, offering a convenient, private, and inclusive space for learning and discussion \cite{fowler2022sex}. At the same time, children's exposure to sexual media content has been found to influence attitudes and contribute to the formation of adversarial sexual beliefs \cite{collins2017sexual}.

\begin{figure}[t]
\centering
\begin{minipage}{.21\textwidth}
  \centering
  \includegraphics[width=0.9\linewidth]{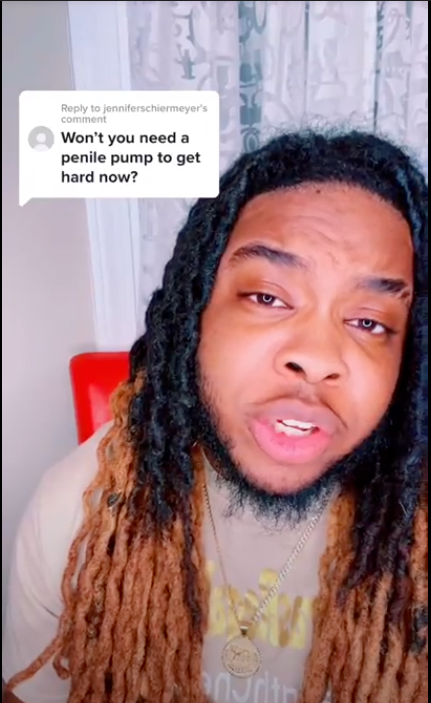}
  \label{fig:test1}
\end{minipage}%
\begin{minipage}{.21\textwidth}
  \centering
  \includegraphics[width=0.9\linewidth]{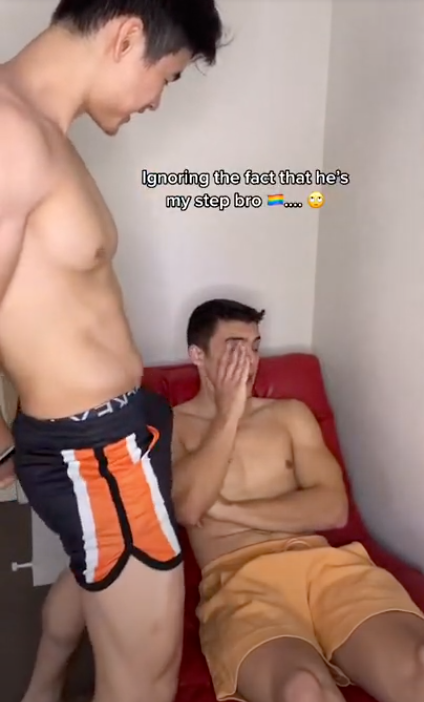}
  \label{fig:test2}
\end{minipage}
\caption{Two screenshots from videos in the dataset. On the left, Nyko (@kingnyko2022) addresses a question about his gender transition. The right is from a sexually suggestive video. }
\label{fig:screenshots}
\end{figure}

\begin{table}
\begin{tabular}{p{0.94\linewidth} }
\hline
\small  (1) \textbf{Educative} {\em (Description)}  \textbf{:} 
\small Video featuring a man discussing a topic while a prominent illustration of a p*n*s with pearly penile papules serves as the background.\\
\hline
\small  (2) \textbf{Suggestive} {\em (Description)} \textbf{:} 
\small Video shows a man holding a pumpkin over his torso while a woman enthusiastically moves her hand inside, exclaiming, "There is so much in there."\\
\hline
\small  (3) \textbf{Educative} {\em (Transcript)}  \textbf{:} 
\small The average banana in the United States is about 5.5 inches long.That's the perfect size for baking banana bread most of the time because ... \\ 
\hline
\small  (4) \textbf{Suggestive} {\em (Transcript)}  \textbf{:} 
\small You are such a good boy. Daddy's so proud of you. \\
\hline
\end{tabular}
\caption{Examples from the dataset, the first two are descriptions, and the latter are video transcripts.}
\label{table:examples}
\end{table}

Unfortunately, efforts to moderate explicit content had unintended consequences, as studies have demonstrated the misidentification of non-explicit content due to flawed algorithms and filtering tech\- niques  \citealp[]{peters2020sexual}. In addition to the above issue, video/video creators (referred to as creators from now on) may also be susceptible to mass reporting. Creators from marginalized communities, partic\- ularly those within the LGBTQIA+ community, face heightened risks of having their educational content wrongfully flagged or removed \footnote{https://mashable.com/article/tiktok-sex-education-content-removal}.

The classification of sexually suggestive and sex education videos presents a complex task, as demonstrated by the examples shown in Table \ref{table:examples}. In example 1, we see that p*n*s illustration is not suggestive, while the video with a man holding a pumpkin in example 2 is suggestive. When we look at the transcripts, we see that in example 3, the creator is talking about myths around p*n*s sizes for pleasurable sex, and in example 4, the audio is suggestive. Considering these complexities, accurately categorizing sexually suggestive and sex education videos necessitates a nuanced understanding of contextual cues, subjectivity, evolving language, and robust algorithmic solutions. 

The contributions of the paper are as follows:

\begin{enumerate}
    \item \textbf{Introduction of SexTok:} A collection of 1000 TikTok videos labeled as Sexually Suggestive, Sex Education, or Others, along with perceived gender expression and transcription.
    \item \textbf{Baselines Evaluation:} We evaluate two transformer-based classifiers as baselines for the task of classifying these videos. Our results indicate that accurately distinguishing between these video types is a learnable yet challenging task.
\end{enumerate}

\subsection*{Trigger Warning: Sexual Content and Explicit Language}
Please be advised that this research paper and its associated content discuss and analyze sexually suggestive and sex education videos. The examples and discussions within this paper may contain explicit or implicit references to sexual acts, body parts, and related topics. The language used may sometimes be explicit. This material is intended for academic and research purposes and is presented to address challenges in content identification and classification.
\section{Related Work}
Automatic detection of sexually explicit videos is an area of active study. In a recent survey, \citealp{cifuentes2022survey} classified the methods into four broad separate strategies. Nudity detection, Analysis of image descriptors ( such as Bag of Visual Words), Motion analysis, and other deep learning techniques. 

Most works around nudity detection are focused on skin-colored region segmentation to identify nudity. This methodology has been extensively explored in the image domain \cite{fleck1996finding}, \cite{wang2005identification} \cite{platzer2014skin}, \cite{garcia2018pornographic},\cite{lee2006implementation}. \cite{ganguly2017detecting}'s work, apart from focusing on the percentage of skin exposure, also gave attention to the body posture of the human in the image and the person's gestures and facial expressions. An alternative strategy is the Bag of Visual Words model,  in which the idea is to minimize the existing semantic gap between the low-level visual features and the high-level concepts about pornography. \cite{deselaers2008bag}, \cite{lopes2009bag}, \cite{ulges2011automatic}, \cite{zhang2013approach}. Approaches based on motion analysis, apart from other features, also capture motion, such as using the periodicity in motion, such as in \cite{4156017}. \cite{zuo2008recognition} uses a Gaussian mixture model (GMM) to recognize porno-sounds, a contour-based image recognition algorithm to detect pornographic imagery, and are combined for the final decision.

Yet still, sexual activity where the human is mostly clothed or has minimal movement is still challenging.
\citealp{peters2020sexual} studied issues surrounding publicly deployed moderation techniques and called for reconsidering how platforms approach this area, especially due to it's high false positive rates and/or low precision rates for certain types of actions.

\section{SexTok Dataset}

This section presents the SexTok dataset \footnote{Data and the experiment codebase will be shared at github.com/enfageorge/SexTok. Videos are shared as links to avoid any potential licensing issues. }, a collection of 1000 TikTok video links accompanied by three key features: Class Label, Gender Expression, and Audio Transcriptions.

\subsection{Terminology and Definitions}

\subsubsection{Class Label}
The first feature, Class Label, is a categorical variable with three possible values: {\em Sexually Suggestive, Sex Education}, and {\em Others}:

{\flushleft \textbf{Sexually Suggestive:}}  This category encompasses videos that purposefully intend to elicit a sexual response from viewers. Determining the presence of sexually suggestive content is subjective.

{\flushleft \textbf{Sex Education:}} This category encompasses videos aimed at enhancing viewers' knowledge, skills, and attitudes concerning sexual and reproductive health. It covers various topics, including but not limited to sexual orientation, gender, and gender-affirming care.

{\flushleft \textbf{Others:}} This category encompasses videos that do not fall within the aforementioned sexually suggestive or sex education categories.

\subsubsection{Gender Expression}
Gender expression is a form of self-expression that refers to how people may express their gender identity \cite{summers2016social}. In this paper, we focus solely on the physical visual cues associated with gender expression. We provide five gender expression labels in the dataset: {\em Feminine, Masculine, Non-conforming, Diverse, and None}.

Feminine and Masculine represent predominantly feminine or masculine expressions, while Non-conforming refers to expressions that deviate from traditional norms. Diverse applies to videos with varying gender expressions among multiple individuals. The None label is for videos without people or only limited visual cues like hands.

The information for the vast majority is not self-reported. When available through the video itself, profile descriptions, or hashtags, we incorporate that information. Otherwise, the annotation is based on the perception of the annotator. {\em This feature is provided only to serve the purpose of evaluating bias in models built on the dataset.}

\begin{table}
\centering
\begin{tabular}{lcccc}
\hline
\textbf{Label} &  \textbf{Train} & \textbf{Val} & \textbf{Test} & \textbf{Total}\\
\hline
\textbf{Sugg} & 140 &	20 & 40 &  200\\
\textbf{Educative} & 140 & 20 & 40 & 200 \\
\textbf{Others} & 420 &	60 & 120 & 600 \\
\hline
\textbf{Total} & 700 &100 &	200 & 1000\\
\hline
\end{tabular}
\caption{Video Distribution by Dataset Split and Class Label.  Sugg: Suggestive, Edu: Educative.The dataset consists mostly of general videos that do not fall into the categories of sexually suggestive or educative. This reflects a more realistic representation of Tiktok's environment.}

\end{table}

\subsection{Dataset Construction}
\subsubsection*{Data Collection}

The data collection process involved the primary annotator creating a new TikTok account and interacting with the platform in various ways to collect the video links. They
carefully watched and hand-selected videos. Two important considerations were taken into account during the dataset construction process: (a) Limit a maximum of five videos per creator in the dataset. (b) Creators appearing in one split of the dataset (train, validation, or test) were excluded from all other splits to ensure independence and prevent data leakage. Detailed information regarding the specific methods used, as well as limitations and ethical considerations, can be found in Appendix \ref{appendix:a}.

\subsubsection*{Annotator Agreement}

A 10\% sample of the dataset was independently annotated by a second author to ensure reliability. Cohen's Kappa scores \citep{cohen1960coefficient} were used to assess annotator agreement. For Gender Expression, the Kappa score was 0.89, indicating substantial agreement. For Class Label, the Kappa score was 0.93, indicating high agreement. These scores validate the consistency and quality of the dataset's annotations.

\subsubsection*{Data Processing: Video download and Audio transcription}

The videos were downloaded without the TikTok watermark using a TikTok downloader.\footnote{\url{https://github.com/anga83/tiktok-downloader}}. The watermark was removed  to reduce unnecessary noise in the data.  

A smaller sample of videos was first transcribed using OpenAI’s whisper (medium) \citep{radford2022robust} and was manually checked for accuracy.  The transcriptions were mostly perfect, with a word error rate of 1.79\%. After this, all the videos were automatically transcribed using Open AI's Whisper (medium).  

\subsection{Dataset Properties}
\begin{small}
\begin{table}

\centering
\begin{tabular}{lccccc}
\hline
\textbf{Label} &  \textbf{Fem} & \textbf{Masc} & \textbf{NC} & \textbf{D} & \textbf{None}\\
\hline
\textbf{Sugg} & 115 & 84	& 0	& 1 & 0\\
\textbf{Edu} & 85 & 84 & 6 & 8 & 17\\
\textbf{Others} & 164 & 170 & 12 & 113 & 141\\
\hline
\textbf{Total} & 364 & 338 & 18 & 122 & 158\\
\hline
\end{tabular}
\caption{Video Distribution by Class Label and Gender Expression. Fem: Feminine, Masc: Masculine, NC: Non-conforming, D: Diverse. Sugg: Suggestive, Edu: Educative. The dataset is predominantly feminine in the suggestive category, while in the educative and others categories, both feminine and masculine gender expressions are relatively balanced and dominant.}

\end{table}
\end{small}

In this section, we provide some general statistics about the SexTok dataset. The dataset comprises 1000 TikTok video links with three features: Class Label, Gender Expression, and Audio Transcriptions. A breakdown by label and dataset split is given in Table 1. A separate breakdown by Gender Expression and dataset split is given in Table 2.

When the audio was transcribed, a percentage of videos were found not to have any text in the audio transcription, specifically → Suggestive - 15.85\%, Educative - 3.97\%, Others - 8.4\%. 

\begin{table*}[ht]

\centering
\begin{small}
\adjustbox{max width=\textwidth}{%
\begin{tabular}{lccccccc}
\hline
\textbf{Group} & \textbf{Acc} & \multicolumn{3}{c}{\textbf{Micro}} & \multicolumn{3}{c}{\textbf{Macro}} \\
\cmidrule(lr){3-5} \cmidrule(lr){6-8} 
 & & P & R & F1 & P & R & F1\\
\hline

Majority & 0.60  & 0.00 & 0.00 & 0.00 & 0.20 & 0.33 & 0.25\\
All Text & 0.68 \footnotesize $\pm$ 0.06 & 0.76 \footnotesize $\pm$ 0.06 & 0.50 \footnotesize $\pm$ 0.06 & 0.60 \footnotesize $\pm$ 0.04 & 0.71 \footnotesize $\pm$ 0.06 & 0.63 \footnotesize $\pm$ 0.03 & 0.64 \footnotesize $\pm$ 0.04\\
Non-empty Text  & 0.75 \footnotesize $\pm$ 0.02 & 0.78 \footnotesize $\pm$ 0.07	& 0.54 \footnotesize $\pm$ 0.02	& 0.64 \footnotesize $\pm$ 0.02  & 0.74 \footnotesize $\pm$ 0.04 &	0.65 \footnotesize $\pm$ 0.01 & 0.68 \footnotesize $\pm$ 0.00\\
Video & 0.70 \footnotesize $\pm$ 0.04 & 0.61 \footnotesize $\pm$ 0.11 & 0.51 \footnotesize $\pm$ 0.07 & 0.55 \footnotesize $\pm$ 0.05 & 0.68 \footnotesize $\pm$ 0.06 & 0.57\footnotesize $\pm$ 0.07 & 0.61 \footnotesize $\pm$ 0.01 \\
\hline
\end{tabular}
}
\end{small}
\caption{\label{table: Results-1} We present the average and standard deviation of results from three different runs of our experiments. We use accuracy, micro-precision, recall, and F1 (with "Others" as a negative class, not included in the scores) and macro-precision, recall, and F1 as metrics. Text-based classification, when transcript is present, has higher overall performance. }

\end{table*}

We also observe that suggestive videos tend to be shorter (median duration: 7.86 secs), and have shorter audio transcriptions (median number of words: 14 words), compared to educative videos that are longer (median duration: 50.80 secs) and have longer audio transcriptions (median number of words: 171.5 words). Detailed dataset video length and transcription length are given in Appendix \ref{appendix:a}.)

\section{Experimental Setups}

In this section, we evaluate the performance of pre-trained transformer-based models on the SexTok dataset to assess its significance. The experiments are divided into two subsections: text classification using video transcripts and video classification. 

For both transformer-based setups, we utilized models downloaded from Hugging Face Transformers \citep{wolf-etal-2020-transformers}, initializing them with three random numbers. Details on hyperparameters are in Appendix \ref{appendix:c}. The reported results are the average of three runs. To assess the performance, we employed four sets of metrics: (1) accuracy, (2) micro precision, recall, and F1 (excluding Others as a negative class from the scores), (3) macro precision, recall, and F1, and (4) overall F1 for each class.

\subsection*{Text Classification using Video Transcript}

We fine-tuned {\tt bert-base-multilingual-cased} \citep{DBLP:journals/corr/abs-1810-04805} to perform text classification on the video transcripts. Since we observed that a small percentage of videos do not yield any text in their transcription, we experimented with two setups. One with all video transcriptions and the other with non-empty transcriptions. 

\subsection*{Video Classification}

We fine-tuned {\tt MCG-NJU/videomae-base}, a VideoMAE base model \cite{videomae} for video classification. The image clips were randomly sampled and preprocessed to align with the default configurations of the model.

\begin{table}
\centering
\begin{small}
\resizebox{0.48\textwidth}{!}{
\begin{tabular}{llll}
\hline
\textbf{Group} & \textbf{Suggestive} & \textbf{Educative} & \textbf{Others}  \\
\hline
Majority & 0.00  & 0.00 & 0.60\\
All Text &  0.30 \footnotesize $\pm$ 0.14 & 0.83 \footnotesize $\pm$ 0.01 & 0.80 \footnotesize $\pm$ 0.02 \\
Non-empty Text  & 0.38 \footnotesize $\pm$ 0.03 & \textbf{0.84} \footnotesize $\pm$ 0.01 & \textbf{0.81} \footnotesize $\pm$ 0.02 \\
Video & \textbf{0.55} \footnotesize $\pm$ 0.02 & 0.63 \footnotesize $\pm$ 0.13  & 0.72 \footnotesize $\pm$ 0.15\\

\hline
\end{tabular}
}
\end{small}
\caption{\label{table: Results-2} We present the overall F1 of each class label with the average and standard deviation of three random runs. Text-based classification gives a higher F1 for educative content when transcription is present, but suggestive content is detected best in videos where educative content is misclassified higher.}
\end{table}

\section{Results and Error Analysis}
\label{Discussion}

The average performance and standard deviation of the models are presented in Tables \ref{table: Results-1} and \ref{table: Results-2}. Based on these results, we draw the following observations:

\begin{itemize}
    \item The most accurate model is the text classifier that evaluated videos with a transcription (75\%). It demonstrates relatively better performance in identifying educative content but often struggles to differentiate between suggestive content and others, and vice versa. However, it should be noted that this implementation is not realistic in a real-world scenario, as TikTok videos can vary in terms of sound presence and spoken language.
    \item Both text-based classifiers exhibit higher F1 scores than the video classifier for the Educative and Others classes. But their performance in detecting suggestive content is is comparatively lower than that of the video classifier.
    \item Notably, neither of the text-based classifiers misclassifies suggestive content as educative, or vice versa, as evident from the confusion matrices in Appendix \ref{appendix:d}. 
    \item The video classifier achieves the highest F1 score for the Suggestive class. However, it frequently confuses Educative and Other videos with each other.
\end{itemize}

To further understand the hard examples for the model, we manually categorized the errors in both text and video classification experiment setups. 

We analysed 54 errors in text classification model. If more than one option was applicable, the video was counted in both: (a) {\em Audio unrelated to class label (50.00\%)}: The audio in these videos consisted of popular songs or speeches that did not contain any words typically associated with the class label. (b) {\em Context clues and Euphemism (25.07\%) }: These videos relied on context clues or employed euphemistic language (9.26\%) or required audio analysis considering the tone and intonation to predict the class label (14.81\%). (c) {\em No or partial transcription (14.81\%)}: Approximately 9.26\% of the videos had no audio that could be transcribed, while 5.56\% had only partial transcriptions available. 
We analyzed 52 errors in video classification. All educative videos that were classified as others, and vice versa, had the same format that both classes do, i.e., a person looking at the camera speaking. Of the 11 suggestive videos that were not classified correctly, in 63\%  of videos, some or all of the video frames had fully or mostly clothed people featured in the video.  A detailed analysis using Transformers-interpret \ref{appendix:d} \cite{Pierse_Transformers_Interpret_2021} also shows that the text classification shows some signs of overfitting to text. 

\section{Discussion}

The results highlight the complexity of accurately identifying sexually suggestive and educative videos on platforms like TikTok. While the results indicate that text analysis can contribute to detecting educative videos, music clips unrelated to the video topic are commonly used, making reliance on transcription alone insufficient. While existing work in pornographic content detection primarily focuses on visual analysis, our results indicate the need for a multi-modal approach since detecting sexual content requires  a more comprehensive understanding encompassing  multiple senses, including audio, speech, and text. 

Addressing these challenges is crucial for developing effective content moderation systems, ensuring appropriate access to sex education, and creating a safer and more inclusive online environment. It is also crucial to be mindful of potential gender expression bias commonly found in visual datasets \cite{meister2022gender}. Moreover, for tasks like this, developing scalable solutions suitable for large-scale systems with millions of users is crucial for effective implementation. Further exploration and investigation of these aspects are left for future research and development.

\section{Conclusion}

This paper introduces a novel task of identifying sexually suggestive and sex-educative videos and presents SexTok, a multi-modal dataset for this purpose. The dataset includes video links labeled for sexual suggestiveness, sex-educational content, and an other category, along with gender expression and audio transcription. The results highlight the challenging and multi-modal nature of the task and suggest that while the dataset is meaningful and the task is learnable,  it remains a challenging problem that deserves future research. This work contributes to promoting online safety and a balanced digital environment.

\section{Acknowledgement}

This work was partially funded by the LGBTQ+ Grad Student Research Funds by The Institute for LGBTQ Studies at the University Of Arizona. We deeply appreciate the invaluable contributions of Shreya Nupur Shakya throughout this work.

\section*{Limitations}
We address the limitations of the SexTok dataset and the accompanying experiments here.
\subsection*{SexTok Dataset}
\begin{itemize}
\item The TikTok account was created and used from a specific geographic location (which will be disclosed in the final version if accepted). This is important to note since the content recommendation of TikTok is influenced by geographic location,\footnote{\url{https://support.tiktok.com/en/account-and-privacy/account-privacy-settings/location-services-on-tiktok}} among other things; hence a geographic bias may be expected, i.e., certain demographics may be more represented than others, especially in terms of languages used, race, ethnicity, etc.
\item The data gathered only represents a small sample of the content available on TikTok and may not represent the entire population of TikTok users or videos.
\item Sexual suggestiveness is treated as a discrete class label in the project, whereas in the real world, it has two important properties. 1) The perception  of what is sexually suggestive may vary depending on the individual's sexual orientation, worldview, culture, location, and experiences and is highly subjective. 2) Some are more suggestive than others, and we do not account for the variation in the strength of suggestiveness here.  
\item The dataset is a small snapshot of the TikTok videos from October 2022 to January 2023. Patterns, slang, and other cues may change over time.
\item Gender expression has many variations but is referred to as discrete labels here, but in real life, it is not. Additionally, this is as perceived by one annotator and, for the majority, not self-reported by the person in the video. Additional expert annotators may be needed to strengthen the confidence in the label.  
\item Despite best efforts, it may be possible that the same creator appears more than five times. This is because creators often create multiple accounts to serve as a backup in case TikTok takes down the original account. This is observed to be increasingly common in the sexually-suggestive and sex-ed domains. We show an example in Figure \ref{same_creator}

Other details : 

The audio content of the TikTok videos comprises various elements, including background music, spoken dialogue (not necessarily from the video creator), or a combination of both. Notably, TikTok provides voice effects that enable users to modify their voices using predefined options.

\subsection*{Experiments}
\item The audio transcription of the videos was created automatically using Open AI's Whisper-medium \citep{radford2022robust}. Hence this is subject to errors, which may impact the performance of the models.
\item For training the models, GPU computing power was used.
\begin{figure}
\centering
\begin{minipage}{.45\textwidth}
  \centering
  \includegraphics[width=1\linewidth]{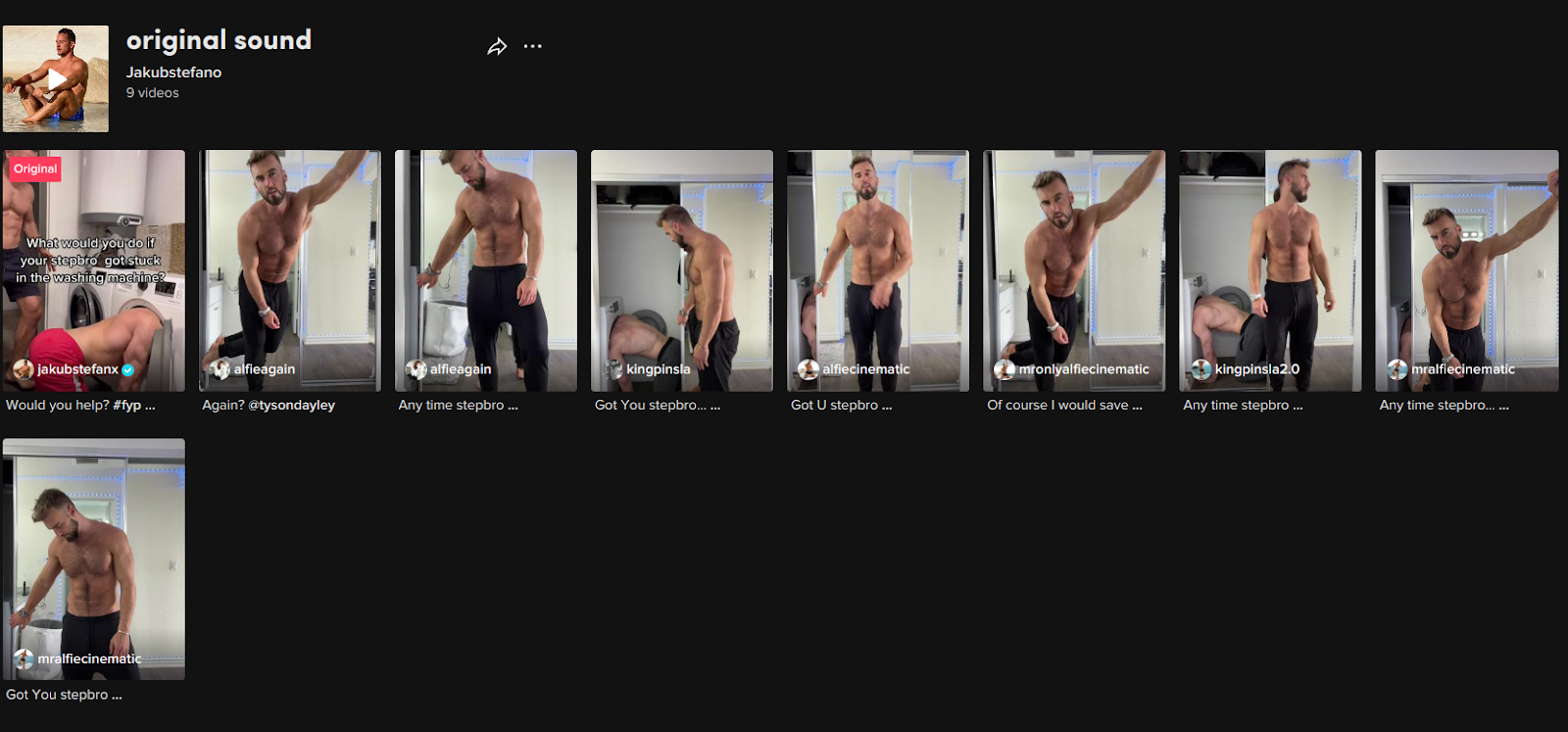}
\end{minipage}%
\caption{This is a partial screenshot from an audio profile page on Tiktok. Each rectangle is a cover image of a video that uses the same audio. The text on the bottom left of each video is the username of the creator of that video. We can see that the same person has multiple accounts posting the same video. }
\label{same_creator}
\end{figure}

\end{itemize}

\section*{Ethics Statement}
We address the ethical considerations and consequences of the SexTok dataset and the accompanying experiments here.
\begin{itemize}
    \item The study's focus is on the technical aspects of the problem. It does not address the broader societal and ethical implications of censorship and of regulating sexually suggestive content on social media platforms. The work only aims to detect sexually suggestive content and sex education content against other video topics but makes no stand on censorship or content regulation of sexually suggestive videos. 
    \item Sexual suggestiveness, as well as perceived gender expression, is a subjective matter and is hence susceptible to annotators' bias.
    \item Gender expression, specifically visual cues only, was annotated and offered only to evaluate bias based on visual cues since such biases are known to exist within large-scale visual datasets \cite{meister2022gender}. The authors do not condone the practice of assigning gender identity based on a person's external appearance since gender is an internal sense of identity \cite{american2015guidelines}. This dataset is not intended to be used for any such practices.
    \item Due to the nature of the problem, and potential licensing issues, the publicly-collected data is not anonymized. 
\end{itemize}

\bibliography{custom}
\bibliographystyle{acl_natbib}

\appendix

\section{Details of Methods used to collect videos}

For sexually suggestive and sex education videos, the annotator interacted with the platform to collect the data in many ways, including search (hashtags, names of people), people reusing the same audio, stitches, duets, the public liked videos of certain profiles pages and the ``For you'' page. Any video that did not appear to belong to either sexually suggestive or sex education was collected and labeled as Others. 

\label{appendix:a}
\subsection{Sexually Suggestive and Sex ed Videos Videos }
\begin{itemize}
\item \textbf{Search :} Hashtags ( including slang usages like \#spicyaccountant), Phrases, and Names of popular creators in a domain (discovered through blogs that talk on the subject).
\item \textbf{Audio Sharing:} TikTok offers multiple people to share and reuse the same audio. So, when a video is found to be, say, sexually suggestive, new creators were discovered by looking into who else used this audio for their video.
\item \textbf{Stitches and Duets:} A \textbf{Duet} allows one creator to post their video side-by-side with a video from another creator on TikTok. A duet contains two videos on a split screen that play at the same time.A \textbf{Stitch} is a creation tool on Tiktok that allows a creator to combine another video on TikTok with the one they are creating. Certain videos added in the dataset were discovered as stitches or duets with another creator.
\item \textbf{Public liked videos:} It is possible to see all videos a certain profile likes by visiting that tab on their profile. By default, this is private but can be set to public. Some profiles share videos of a topic by redirecting visitors to their liked videos. Many videos were found and added to the dataset through this method.
\item \textbf{"For you" Page:} It's a recommended feed of videos from creators the user might not follow. The annotator liked and saved videos of sexually suggestive nature, so some similar videos were recommended on the For you Page.
\end{itemize}

\subsection{Other Videos}
There are three main strategies for collecting these videos. 
\begin{itemize}
    \item Videos that appeared on the TikTok home page when no user was logged in
    \item Videos shared with \#learnontiktok hashtag
    \item Videos that reused audio that was also used in a sexually suggestive video.
\end{itemize}
Each makes up one-third of the total videos collected.

\section{Detailed stats for transcript length and video length}

\begin{table}[H]
\centering
\adjustbox{max width=\textwidth}{%
\begin{tabular}{lcccc}
\hline
\textbf{Parameter} &  \textbf{Sugg} & \textbf{Edu} & \textbf{Others} & \textbf{Total}  \\
\hline
\textbf{Mean} &  \small 16.46 &   \small 231.18  & \small 82.18 &  \small 98.83 \\
\textbf{Median}  & \small 14.00 & \small 171.50 & \small 31.00	 &  \small 33.00  \\
\textbf{Std} &    \small 14.33 & \small 220.81 & \small 126.37 & 	\small 156.08 \\
\hline
\end{tabular}
}
\caption{Mean, Median, and Standard Deviation of words present in video transcripts. Words were tokenized using the NLTK package.  Sugg stands for Suggestive, and Edu stands for educative. Suggestive videos tend to be significantly shorter than the other classes.}

\end{table}

\begin{table}[H]
\centering
\begin{tabular}{lcccc}
\hline
\textbf{Parameter} &  \textbf{Sugg} & \textbf{Edu} & \textbf{Others} & \textbf{Total}  \\
\hline
\textbf{Mean} & 8.96 &  66.41 &  39.99 & 39.06  \\
\textbf{Median} & 7.86 &  50.80 & 	28.30 & 23.16   \\
\textbf{Std} & 3.82 & 56.92 & 37.88 &	42.90 \\
\hline
\end{tabular}
\caption{Mean, Median, and Standard Deviation of videos in the dataset in seconds. Sugg stands for Suggestive, and Edu stands for educative. Suggestive videos tend to be significantly shorter than the other classes.}

\end{table}

\section{Hyperparameters}

Hyperparameters not mentioned below, are default values from Huggingface.
\label{appendix:c}

\begin{table}[H]
\centering
\begin{tabular}{l l}
\hline
 \textbf{Parameter} &  \textbf{Value} \\
 \hline
 Batch size & 16 \\
 Initial Learning Rate  & 1e-5 \\
 Weight Decay & 0.01 \\
 Warmup Ratio & 0.1 \\
 Learning Rate Optimiser & AdamW \\
\hline
\end{tabular}
\caption{Hyperparameters used for the Text Classification Task}
\end{table}

\begin{table}[H]
\centering
\begin{tabular}{lc}
\hline
 \textbf{Parameter} &  \textbf{Value} \\
 \hline
 Batch size & 8 \\
 Initial Learning Rate  & 5e-5 \\
 Warmup Ratio & 0.1 \\
 Learning Rate Optimiser & AdamW \\
\hline
\end{tabular}
\caption{Hyperparameters used for the Video Classification Task}
\end{table}
\label{appendix:d}
\section{Transformer Interpret}
Refer to Figure 3 on the next page.
\newpage
\begin{figure*}[!htb]]
\centering
\includegraphics[width=0.9\textwidth]{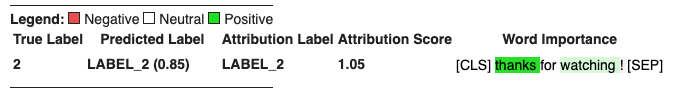}
\includegraphics[width=0.9\textwidth]{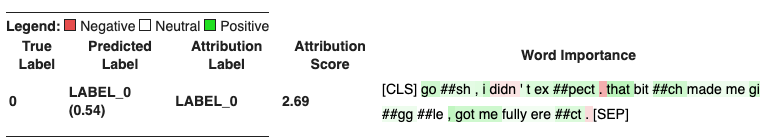}
\includegraphics[width=0.9\textwidth]{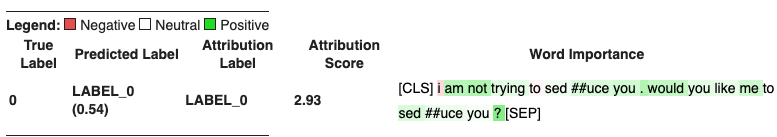}
\caption{Three example transcription and its predictions explanation  visualized using Transformers Interpret, a model explainability tool.}
\end{figure*}
\end{document}